\pdfoutput=1
\documentclass[10pt,twocolumn,letterpaper]{article}
\usepackage{iccv}
\usepackage{times}
\usepackage{epsfig}
\usepackage{graphicx}
\usepackage{amsmath}
\usepackage{amssymb}
\usepackage{dsfont}
\usepackage{multirow}
\usepackage{multicol}
\usepackage{graphics}
\usepackage{subfigure}
\usepackage{xcolor}
\usepackage{url}
\usepackage{float}
\usepackage[accsupp]{axessibility}  % Improves PDF readability for those with disabilities.

\definecolor{darkgreen}{RGB}{40, 133, 65}

\usepackage[breaklinks=true,bookmarks=false]{hyperref}

\iccvfinalcopy % *** Uncomment this line for the final submission

\def\iccvPaperID{10} % *** Enter the ICCV Paper ID here

\ificcvfinal\fi

\makeatletter
\newcommand{\multiline}[1]{%
  \begin{tabularx}{\dimexpr\linewidth-\ALG@thistlm}[t]{@{}X@{}}
    #1
  \end{tabularx}
}
\makeatletter
\def\hlinewd#1{%
  \noalign{\ifnum0=`}\fi\hrule \@height #1 \futurelet
  \reserved@a\@xhline}
  
\begin{document}

%%%%%%%%% TITLE
\title{Probeable DARTS with Application to Computational Pathology}

\author{Sheyang Tang$^{1}$\thanks{Equal contribution}, Mahdi S. Hosseini$^{2}$\footnotemark[1], Lina Chen$^{3}$, Sonal Varma$^{4}$, Corwyn Rowsell $^{5}$, \\
Savvas Damaskinos $^6$, Konstantinos N. Plataniotis $^7$, and Zhou Wang$^1$\\

% $^1$The Department of Electrical and Computer Engineering, University of Waterloo \\
% $^2$The Department of Electrical and Computer Engineering, University of New Brunswick \\
% $^3$Sunnybrook Hospital, University of Toronto\\
% $^4$Division of Pathology, St. Michaels Hospital, Toronto, ON, M4N 1X3, Canada\\
% $^5$Department of Pathology and Molecular Medicine, Queen's University \\and Kingston Health Sciences Center\\
% $^6$Huron Digital Pathology, St. Jacobs, ON, N0B 2N0, Canada\\
% $^7$The Edward S. Rogers Sr. Department of Electrical \& Computer Engineering, University of Toronto\\
% \tt\small\color{purple}\url{https://github.com/mahdihosseini/DARTS-ADP}
$^1$University of Waterloo, Canada~~~$^2$University of New Brunswick, Canada~~~\\
$^3$Sunnybrook health science center, Canada~~~$^4$Kingston Health Sciences Center, Canada\\
$^5$St. Michaels Hospital, Canada~~~$^6$Huron Digital Pathology, Canada~~~$^7$University of Toronto, Canada\\
\tt\color{purple}\url{https://github.com/mahdihosseini/DARTS-ADP}
}

\maketitle
\ificcvfinal\thispagestyle{empty}\fi

%%%%%%%%% ABSTRACT
\begin{abstract}
   AI technology has made remarkable achievements in computational pathology (CPath), especially with the help of deep neural networks. However, the network performance is highly related to architecture design, which commonly requires human experts with domain knowledge. In this paper, we combat this challenge with the recent advance in neural architecture search (NAS) to find an optimal network for CPath applications. In particular, we use differentiable architecture search (DARTS) for its efficiency. We first adopt a probing metric to show that the original DARTS lacks proper hyperparameter tuning on the CIFAR dataset, and how the generalization issue can be addressed using an adaptive optimization strategy. We then apply our searching framework on CPath applications by searching for the optimum network architecture on a histological tissue type dataset (ADP). Results show that the searched network outperforms state-of-the-art networks in terms of prediction accuracy and computation complexity. We further conduct extensive experiments to demonstrate the transferability of the searched network to new CPath applications, the robustness against downscaled inputs, as well as the reliability of predictions.
\end{abstract}

%%%%%%%%% BODY TEXT

\section{Introduction}
Recent years have witnessed great advances in AI-based Computational Pathology (CPath) \cite{litjens2017survey, janowczyk2016deep}. The emerging AI techniques have shown their superiority in more accurate, efficient, and large-scale medical diagnoses \cite{bera2019artificial}. In particular, Convolutional Neural Networks (CNNs) have been widely employed to extract meaningful information from medical images for various pathology applications, including disease diagnoses \cite{cirecsan2013mitosis, zhao2017automatic}, medical image segmentation \cite{ronneberger2015u, song2016accurate}, etc. Yet designing the network architectures has long been a manual process that requires adequate domain knowledge. As a result, it has become a common standard that architectures from CV applications (such as ResNet \cite{he2016deep} and GoogLeNet \cite{szegedy2015going}) are transferred for technical developments in other fields, including CPath \cite{schaumberg2017h, wang2016deep}. 

\begin{table}[htp]
    \setlength\tabcolsep{1pt} 
    \center
    \caption{\label{cifar_adp}Comparison between CV and CPath datasets.}
	\label{results_cell_optimizer}
    \scriptsize{
    \begin{tabular}{ c||c|c||c }
        \hlinewd{1pt}
        \multirow{2}{*}{\textbf{General Stats}} &
        \multicolumn{2}{c||}{\textbf{CV}} &
        \multicolumn{1}{c}{\textbf{CPath}} \\
        \cline{2-4}
        & CIFAR-10 & CIFAR-100 & ADP \\
        \hline
        \hline
        Training size & 50000 & 50000 & 14134\\
        Validation size & - & - & 1767 \\
        Test size & 10000 & 10000 & 1767 \\
        Resolution & 32x32 & 32x32 & 272x272 \\
        \# classes & 10 & 100 & 33 \\
        Label form& single-label & single-label & multi-label\\
        Background& various & various & white\\
        \hlinewd{1pt}
    \end{tabular}
    }
\end{table}
\begin{figure}[htp]
\centering
    \subfigure[CIFAR]{\includegraphics[width=0.2\textwidth]{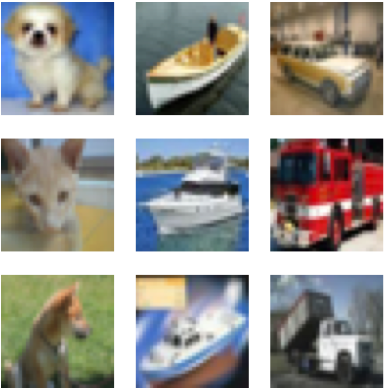}\label{fig:cifar}}
    \subfigure[ADP]{\includegraphics[width=0.2\textwidth]{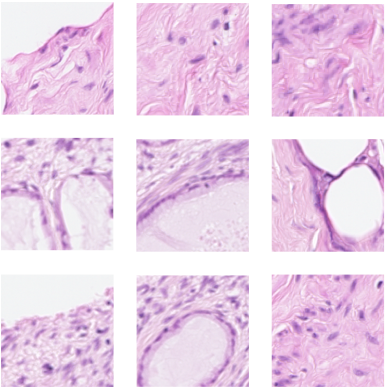}\label{fig:adp}}
\caption{Sample images from CV and CPath datasets.}
\label{fig:sample_images}
\end{figure}

The ultimate question is whether transferring architectures between the two domains is an efficient strategy. To answer this question, we first demonstrate how CV and CPath datasets are different. Here we compare the CIFAR \cite{krizhevsky2009learning} and ADP \cite{hosseini2019atlas} datasets. Besides different data structures shown in Table.~\ref{cifar_adp}, the nature of images from both sides is also different, which makes CV datasets more complicated. \textit{First}, the pixel resolution in CPath is fixed, corresponding to a fixed field of view (FOV) size. The root cause of such uniformity is the acquisition of whole slide images by a scanner in a much more controlled environment from both optics and illumination viewpoint \cite{hosseini2019atlas}. In contrast, the pixel resolution in CV is randomly distributed across different images due to different image setup and configurations. CV images are captured in natural scenes where the distance has much variance. Examples from each imaging modality are shown in Figure \ref{fig:cifar}, where the ship images are taken from a further distance than the dog ones, resulting in larger pixel size and lower resolution. \textit{Second}, target objects in CV images only occupy part of the whole FOV and the rest are background which is irrelevant to the class label. Note that the diversity of the background in Figure \ref{fig:cifar} is very high. This is quite different in CPath where the background information is obtained from an empty area of the sample using uniform white light illumination \cite{hosseini2019atlas} --leading to more uniform and homogeneous images. This is illustrated in Figure \ref{fig:adp}, where the white part denotes the background. In the light of this difference, we form a hypothesis that such simplified imaging modality in CPath translates to simpler network architecture compared to CV. To this end, new network architectures should be designed for CPath applications. 

Neural architecture search (NAS) has recently been proposed to automate the design of neural networks by searching for the optimal network structure on a given dataset. In many CV applications, NAS has outperformed state-of-the-art manually designed networks in terms of prediction accuracy and computation complexity \cite{elsken2019neural}. In medical image analysis, it has been utilized to find suitable networks for various applications, such as image segmentation for Magnetic Resonance Imaging (MRI) \cite{kim2019scalable, yu2020c2fnas, bae2019resource}, ultrasound imaging \cite{weng2019unet}, disease diagnoses from Computed Tomography (CT) scans \cite{jiang2021learning, he2021automated}, etc. In pathology, however, NAS is not fully explored. There is a lack of a general framework that can be easily extended to various CPath applications.

In this work, we propose an architecture search platform based on differentiable architecture search (DARTS) \cite{liu2018darts}. We choose DARTS because it is gradient-based and thus much more efficient and computation-friendly than other searching strategies including reinforcement learning \cite{zoph2016neural} and evolutionary algorithms \cite{real2019regularized}. DARTS achieves this by relaxing the search space to be continuous and dividing the whole pipeline into a search phase and an evaluation phase. However, in CV applications, it is reported that DARTS tends to exhibit overfitting issues, and the searched architecture does not generalize well in the evaluation phase \cite{liang2019darts+, zela2019understanding}. To combat these challenges, we first conduct searching on CIFAR \cite{krizhevsky2009learning} and utilize a probing metric \textit{stable rank} \cite{hosseini2020adas} for each layer. In this way, we can better monitor the searching process and show that the overfitting issue comes from improper hyperparameter tuning. In addition, we use an adaptive optimizer \textit{Adas} \cite{hosseini2020adas} that automatically tunes the learning rates for each layer based on their probing metrics, so that the generalization ability of the searched architecture is improved. We then apply this searching framework on ADP \cite{hosseini2019atlas}, which contains a great variety of histological tissue types that are representative enough, so that the searched architecture can generalize well in different CPath applications. The searched network outperforms the state-of-the-art architectures in the speed-accuracy trade-off, which is crucial for real-time high-throughput CPath applications. We further conduct extensive experiments to show the transferability of the searched architecture on new CPath datasets, demonstrate its robustness against decreased input images, and verify its superiority in extracting label-pertinent features. Our main contributions are listed below:
\begin{itemize}
\item We use a probing metric to show that the existing DARTS framework lacks proper hyperparameter tuning, and use an adaptive optimizer to improve the generalization ability of the searched model;

\item We apply the proposed searching platform on CPath applications and show the superiority of the searched model in prediction accuracy and computation complexity;

\item We demonstrate the transferability of the searched architecture in various CPath applications, show its robustness against decreased resolutions and its reliability in prediction.
\end{itemize}

\section{Related Works}

\begin{table}[htp]
    \setlength\tabcolsep{1pt} 
    \center
    \caption{\label{tab:related}Summary of NAS applications in medical image analysis.}
    \scriptsize{
    \begin{tabular}{ c|c|c|c }
        \hlinewd{1pt}
        \multirow{2}{*}{\textbf{Task}} &
        \multicolumn{3}{c}{\textbf{Searching Strategy}}\\
        \cline{2-4} 
        & Gradient-based & Reinforcement Learning & Evolutionary Algorithms \\
        \hline
        \hline
        Segmentation & \cite{weng2019unet,kim2019scalable,dong2019neural} & \cite{bae2019resource} & \cite{yu2020c2fnas} \\
        Classification & \cite{peng2020multi,he2021automated} & \cite{hosseini2020transferability} & \cite{jiang2021learning} \\
        \hlinewd{1pt}
    \end{tabular}
    }
\end{table}

As NAS has achieved promising results in many CV applications \cite{elsken2019neural}, several attempts are made to utilize NAS techniques to find optimum architectures for applications in medical image analysis. Based on the task and the searching strategy, these works can be categorized as in Table.~\ref{tab:related}. In applications of image segmentation, most works adopt a U-net structure, where detail configurations are searched in different manners. \cite{weng2019unet,kim2019scalable} use differentiable architecture search to find cell structures as building blocks in the encoder and decoder. Bae et al.~\cite{bae2019resource} utilize reinforcement learning to search for hyper-parameter configurations of the U-Net architecture. Yu et al.~\cite{yu2020c2fnas} first search for cell connections to form a U-Net topology using evolutionary algorithms, and then search for operations within each cell. Dong et al.~\cite{dong2019neural} extend the differentiable searching framework to work in adversarial training.

For classification applications, the searching is more task-specific. Using gradient-based searching, Peng et al.~\cite{peng2020multi} develop a network to predict distant metastases on PET-CT images, and He et al.~\cite{he2021automated} design a network for COVID-19 detection with Chest CT Scans. Hosseini et al.~\cite{hosseini2020transferability} use a reinforcement learning-based controller to find the best parameter configuration of a CNN model for histological tissue type classification. Jiang et al.~\cite{jiang2021learning} search for a network to classify pulmonary nodules with evolutionary algorithms. 

To the best of our knowledge, there hasn't been any work that fully explores the potentials of NAS in digital pathology applications.
%------------------------------------------------------------------------
\section{Proposed Method}
% briefly overview what are we presenting in this section (no more than 3 lines) i.e. purpose of this section and how it's organized

\begin{figure*}[htp]
\centering
\includegraphics[width=0.8\paperwidth]{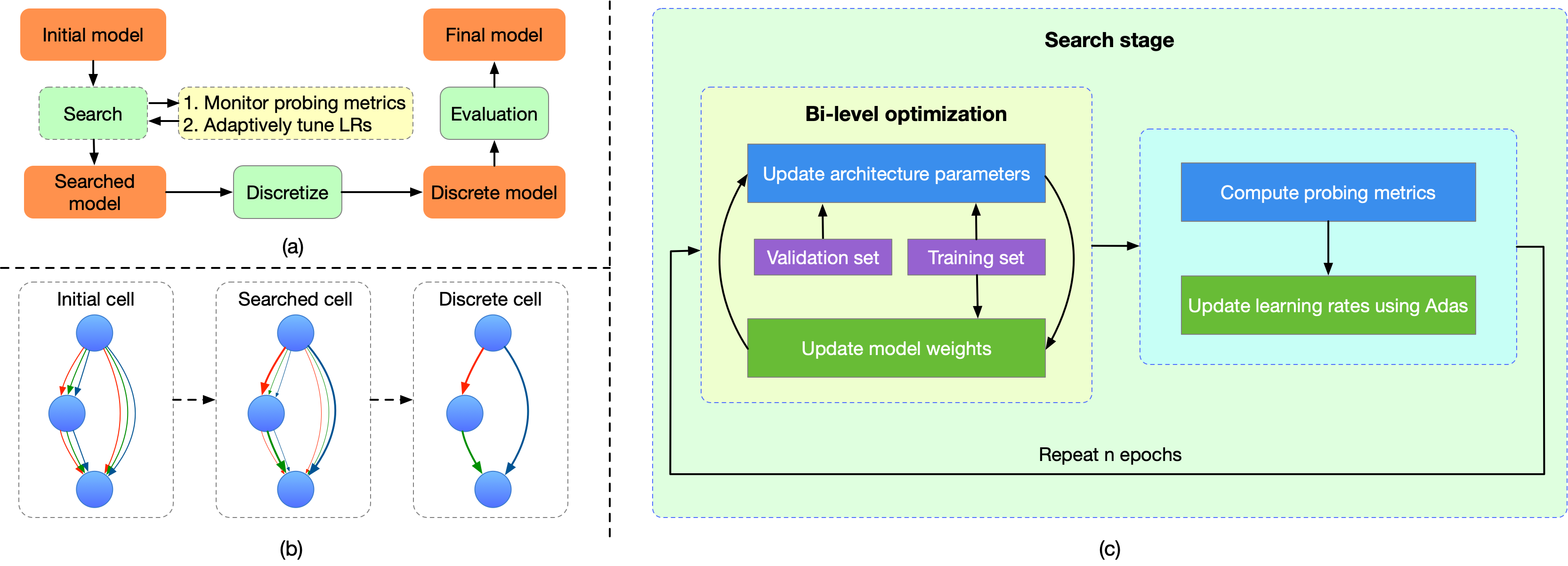}
\caption{(a) Overview of the architecture searching pipeline. The network architecture is searched during the search stage and discretized when finished. It is then trained from scratch for final evaluation. (b) Illustration of the cell structure during searching. (c) Details of the search stage. During each searching epoch, the architecture parameters and model weights are updated alternately through bi-level optimization, then the learning rates for each convolutional layer are tuned based on probing metrics.}
\label{fig:overview}
\end{figure*}

In this section, we introduce our searching algorithm. We first review the basic concepts of DARTS \cite{liu2018darts}, then show how the existing DARTS framework can be improved using a probing metric and a new optimizer. Finally, a network size-based searching is proposed to seek a trade-off between prediction accuracy and model complexity.

\subsection{Review on DARTS}
\label{subsec:darts}
The goal of DARTS \cite{liu2018darts} is to search for two types of cells (namely \textit{normal} and \textit{reduction}) as building blocks, which are stacked to form a full network. Each cell is represented as a directed acyclic graph with $N$ nodes, including two input nodes, intermediate nodes and one output node. Every node $x_{i}$ is a latent representation (\eg, feature map in CNN) and every edge $(i,j)$ is a mixture of weighted candidate operations in a pre-defined operation search space $\mathcal{O}$ (\eg, convolution, skip-connection). The output $\bar{o}_{i,j}$ of an edge $(i,j)$ is then a weighted sum of candidate operations \cite{liu2018darts}:
\begin{equation}
    \bar{o}_{i,j}=\sum_{o\in\mathcal{O}}\frac{\exp\left(\alpha^o_{i,j}\right)}{\sum_{o'\in{\mathcal{O}}}\exp\left(\alpha^{o'}_{i,j}\right)}\cdot{o\left(x_{i}\right)},
\end{equation}
where $\alpha^o_{i,j}$ is an architecture parameter for weighting operation $o\left(x_i\right)$. The output of an intermediate node $x_j$ is the sum of all input edges, \ie, $x_j=\sum_{i<j}\bar{o}_{i,j}\left(x_j\right)$. The output node of a cell is the concatenation of all intermediate nodes. Normal cells keep the input resolution while reduction cells decrease resolutions with stride 2 in all candidate operations.

In the searching procedure, the network weight $w$ and architecture parameter $\alpha$ are jointly learned via bi-level optimization \cite{liu2018darts}:
\begin{equation}
\begin{split}
    &\min_{\alpha}\mathcal{L}_{val}\left(w^*\left(\alpha\right),\alpha\right) \\
    \mathrm{s.t.}\quad &w^*\left(\alpha\right)=\arg\min_w\mathcal{L}_{train}\left(w,\alpha\right),
\end{split}
\end{equation}
where $\mathcal{L}_{val}$ and $\mathcal{L}_{train}$ denote the validation and training datasets, respectively. Using gradient descent, $w$ and $\alpha$ can be updated alternatively during each training iteration.

When the searching is finished, the discrete cell architecture is obtained by replacing each edge by the operation with the largest architecture weight, then selecting the two strongest input edges for each intermediate node. Fig.~\ref{fig:overview} (b) illustrates the evolution of cell structure during searching. The discrete network is then retrained from scratch for final evaluation. The whole process is shown in Fig.~\ref{fig:overview} (a).

\subsection{Explainable Metrics for Probing}
% - explain the metrics
% - show evolution of metrics in figure example
% - make preliminary assessment/conclusion how existing DARTS framework lacks of proper tuning of default HPs. Also show some preliminary results on improving accuracy of searching by tuning the LR.
\label{subsec:metric}
To monitor the searching process of DARTS, we adopt the explainability metric \textit{stable rank} to probe the intermediate convolutional layers in different cells and quantify their learning quality as explained in \cite{hosseini2020adas}. Given a convolutional weight matrix, we first decompose it by low-rank factorization. This factors out the perturbation noise in the layer while keeping the most useful information in the low-rank component. The stable rank $\mathcal{S}$ is the normalized sum of the singular values of the low-rank matrix. It measures the norm energy of the convolutional weights and encodes the low-rank structure’s space span of the output mapping. A higher value indicates better propagation of information through a convolutional layer \cite{hosseini2020adas}.

% Consider the convolutional weight as a 4-dimensional tensor $\mathbf{W_{4D}}\in\RR^{h\times w\times n_i\times n_o}$, where $h$ and $w$ denote the kernel height and width, and $n_i$ and $n_o$ denote the input and output channel sizes, respectively. The tensor $\mathbf{W_{4D}}$ can be unfolded either on the input channel as $\mathbf{W}\in\RR^{whn_o\times n_i}$ or output channel as $\mathbf{W}\in\RR^{whn_i\times n_o}$. In both forms, $\mathbf{W}\in\RR^{m\times n}$, where $n\le m$. The low-rank matrix $\widehat{\mathbf{W}}_l$ is obtained by:
% \begin{align}
% \mathbf{W}\xrightarrow{\text{factorize}}\widehat{\mathbf{W}}_l+E_l,
% \end{align}
% where $E_l$ represents noise perturbation. Then, the stable rank is computed by
% \begin{equation}
%     s\left(\widehat{\mathbf{W}}_l\right)=
%     \frac{1}{n\sigma_1^2\left(\widehat{\mathbf{W}}_l\right)}
%     \sum_{i=1}^{n'} \sigma_i^2\left(\widehat{\mathbf{W}}_l\right),
% \end{equation}
% where $n'=\mathrm{rank}\left\{\widehat{\mathbf{W}}_l\right\}$, and $\sigma_1\geq\sigma_2\geq...\geq\sigma_{n'}$ are the singular values of $\widehat{\mathbf{W}}_l$. The stable rank measures the normalized low-rank energy of convolutional weights, and encodes the low-rank structure’s space span of the output mapping. Higher value indicates better propagation of information through a convolutinal layer.

\begin{figure}[htp]
\centering
\includegraphics[width=0.47\textwidth]{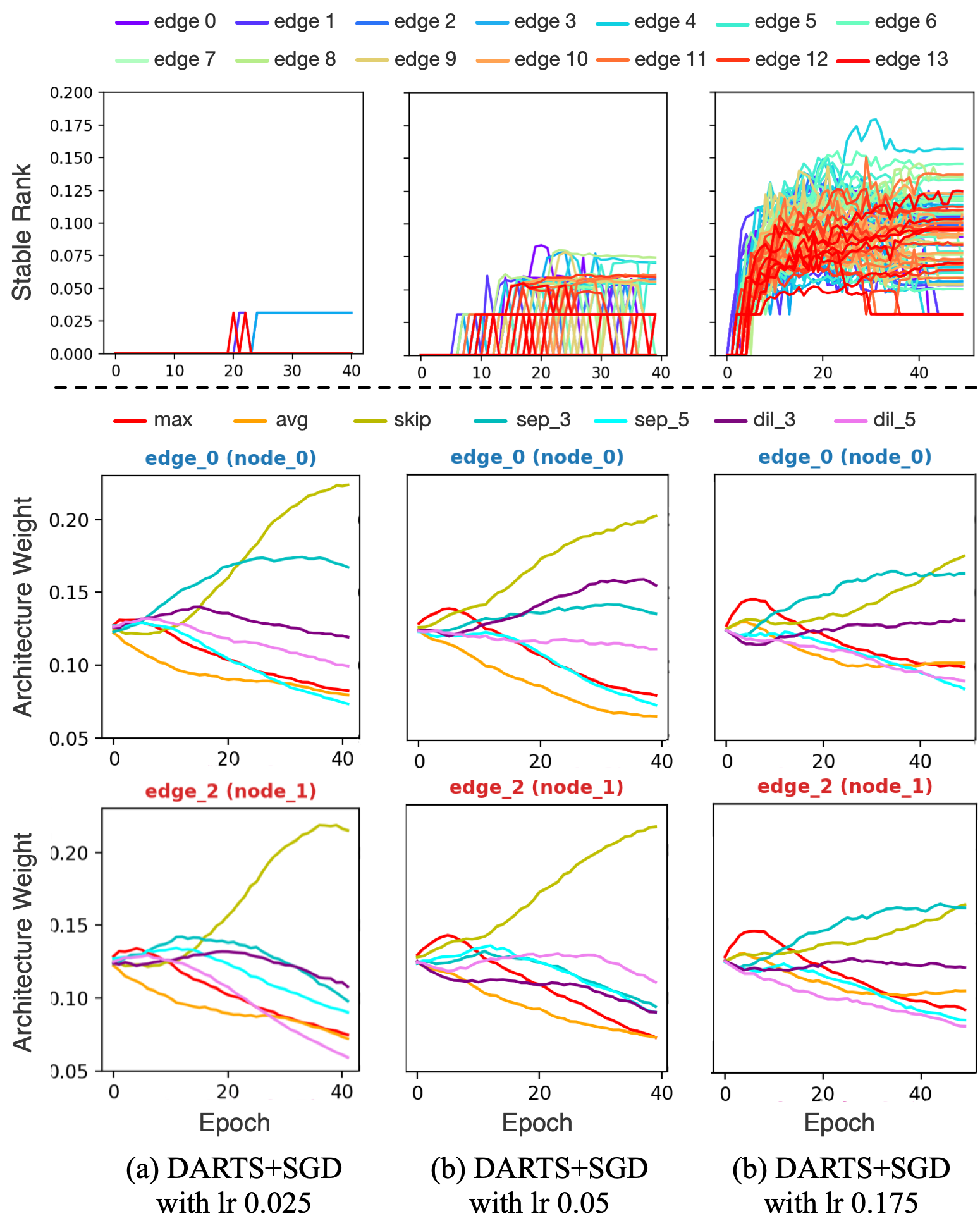}
\caption{Comparison of the stable rank and architecture weight evolution on CIFAR100.}
\label{fig:kg_weight_compare}
\end{figure}

Using this probing metric, we monitor the searching phase of the existing DARTS framework applied on the CIFAR100 dataset \cite{krizhevsky2009learning}. The left column of Fig.~\ref{fig:kg_weight_compare} shows an example of the stable rank evolution (top row) of the layers in one cell as well as the architecture weights evolution (bottom 2 rows) in two edges. We can see from the stable rank evolution that the convolutional layers are not learning well in the original DARTS with stochastic gradient descent (DARTS+SGD) and default initial learning rate 0.025 \cite{liu2018darts} -- most layers generate zero stable rank through all training epochs. Recall the edge structure introduced in Sec.~\ref{subsec:darts}, each candidate operation is multiplied by a weight, which is between $0$ and $1$. This makes their gradients small during backpropagation. Therefore the convolutional layers are learning slowly. The bottom two images show that skip-connections (green curves) are preferred. The same phenomenon is reported in \cite{liang2019darts+} that when searched on CV datasets, the original DARTS tends to select too many skip-connections, which is a kind of overfitting, resulting in a shallow network with poor representation ability.

In light of this, we increase the initial learning rate from $0.025$ to $0.05$ and $0.175$. The stable rank evolution of layers in the same cell, as well as the architecture weights in the same edges, are shown in Fig.~\ref{fig:kg_weight_compare} (b) and (c). With the increase in initial learning rates, more layers generate a higher stable rank, which means they're learning better. In the meantime, the architecture weights evolution reveals that the preference for skip-connections is suppressed. With the help of the probing metric, we know that the existing DARTS framework lacks proper tuning in hyperparameters and how the searching can be improved.

\subsection{Improving DARTS Performance}
% - now we have showed (in previous subsection) that the metrics can be sued to tune proper HPs, then why not utilized optimizer that incorporates this metric for LR adjustment for searching and that is Adas.
% show the architecture weight evolution and show how it gets better by incorporating better optimizer for the searching; perhaps by not falling into bias selection of skip, etc.
\label{optimizer}

\begin{figure}[htp]
\centering
\includegraphics[width=0.3\paperwidth]{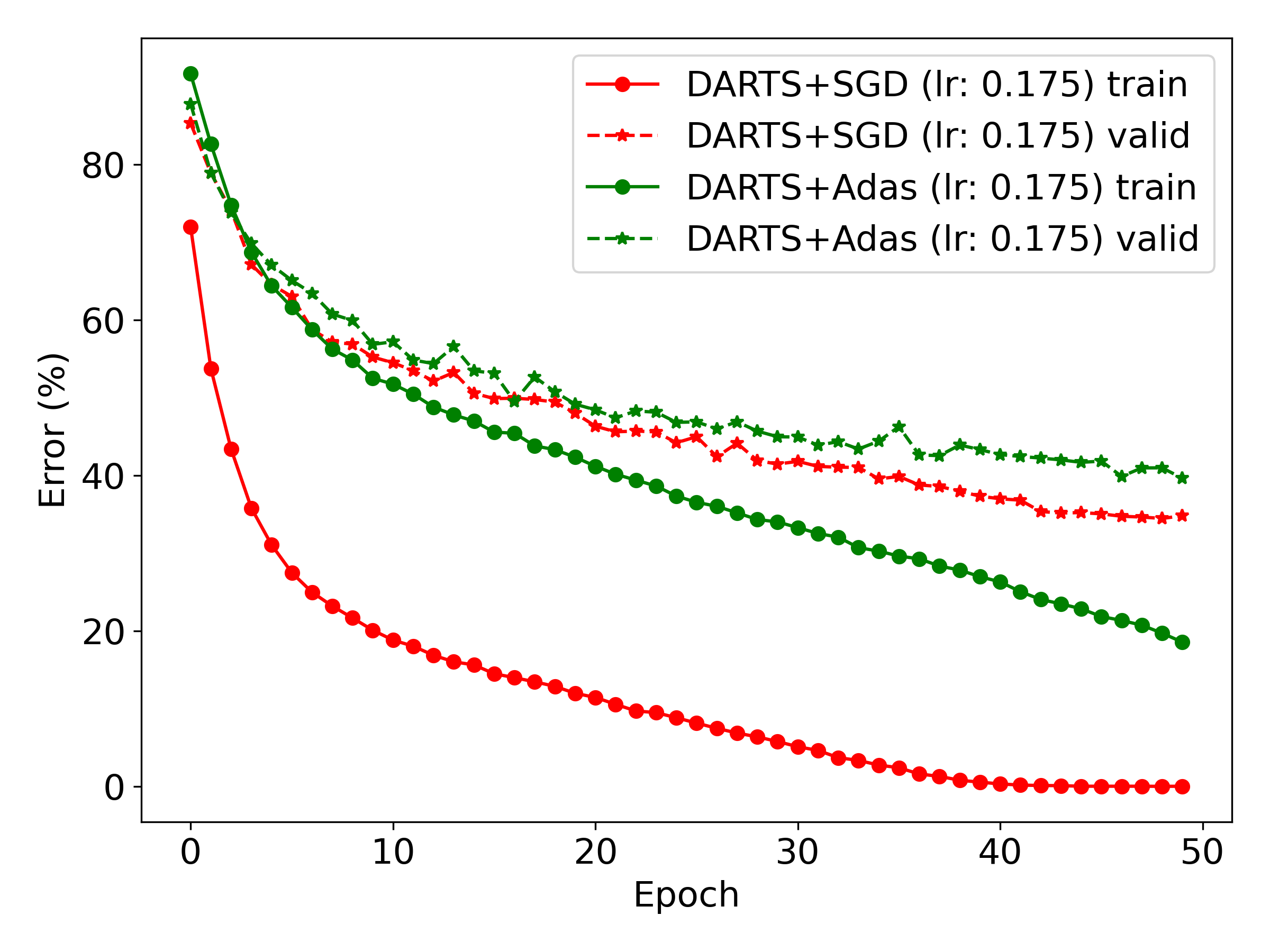}
\caption{Training and validation error during searching on CIFAR-100.}
\label{fig:train_valid}
\end{figure}

We have shown in the previous section that the probing metric can be used to tune proper initial learning rates for DARTS. Then why not utilize an optimizer that incorporates such metric for learning rates adjustment during searching? To verify this, we adopt the Adas optimizer \cite{hosseini2020adas}, which adaptively adjusts the learning rate for each layer based on their stable rank evolution. At the end of each searching epoch, it first computes the difference in stable rank over consecutive epochs for each convolutional layer, and then adds (with a weight) result to the learning rate momentum. A hyperparameter \textit{scheduler beta} is used for weighting this term. This process is illustrated in Fig.~\ref{fig:overview} (c). The Adas optimizer is aware of the learning quality of each layer and therefore tunes their learning rates accordingly. 

Fig.~\ref{fig:train_valid} shows the training and validation errors when searching with different optimizers and initial learning rates. We can see that SGD with a 0.175 initial learning rate leads to overfitting during searching. The final gap between training and validation error is around $40\%$. While for Adas, the overfitting problem is reduced. The resulting final gap is around $20\%$. This indicates that Adas improves the searching process of DARTS with better generalization ability.

\subsection{Network Size-based Searching}
\label{subsec:size}
% - show the finalized architecture for CIFAR
% - show the finalized architecture for ADP (we need to tune the number of cell)
We investigate the trade-off between network size and test performance by searching for the optimal architectures with different numbers of cells and intermediate nodes. This trade-off is crucial for high-throughput CPath applications in real-time. On the ADP dataset, DARTS+Adas obtains the best-performing architecture. It consists of four cells, and each cell contains three intermediate nodes. Fig.~\ref{fig:snapshots_nodes} (c) and (d) show the snapshots of the cell structures.

% \begin{figure}[htp]
% \centering
%     \subfigure[Normal cell]{\includegraphics[width=0.47\textwidth]{figures/normal_n_3_c_4.eps}}
%     \vfill
%     \subfigure[Reduction cell]{\includegraphics[width=0.43\textwidth]{figures/reduction_n_3_c_4.eps}}
% \caption{Best cells found in ADP.}
% \label{fig:best_adp}
% \end{figure}

%------------------------------------------------------------------------
\section{Experiments and Results}
Our experiments contain two stages. In the first stage, we search for the optimum architectures on CIFAR and ADP. In the second stage, we evaluate the transferability of the architecture searched on ADP, as well as its robustness and reliability in various cases.
\subsection{Experimental Setup}
\label{subsec:setup}

\section{Architecture Search}
We carry out the searching on CIFAR and ADP datasets. The search space of candidate operations is the same as in \cite{liu2018darts}, including 1) 3x3 separable convolution, 2) 5x5 separable convolution, 3) 3x3 dilated separable convolution, 4) 5x5 dilated separable convolution, 5) 3x3 max pooling, 6) 3x3 average pooling, 7) skip connection, and 8) zero operation. We stack the cells sequentially to build a network for searching and evaluation. The details of network structures can be found in Section Network Structures of the supplementary material. In both CIFAR and ADP experiments, we test two optimization strategies for optimizing model weights, \ie, DARTS+SGD and DARTS+Adas. Detailed setup can be found in Section Hyperparameters of the Supplementary Material.

\section{Architecture Evaluation}
The searched network is discretized and then trained from scratch for final evaluation. Following \cite{liu2018darts}, in each parameter setting, we conduct four independent runs of searching with different random seeds. We then perform a quick evaluation for each searched architecture by training them from scratch for 100 epochs and pick the best-performing one. The finalized architecture is trained from scratch for 600 epochs in three independent runs. We report the means and standard deviations of test accuracy. Training details can be found in Section Hyperparameters of the Supplementary Material.
% \subsubsection{Architecture Search} 
% In CIFAR experiments, we train the network for 50 epochs with batch size 64 and initial channels 16. We test two optimization strategies for optimizing model weights, \ie, SGD and Adas. For DARTS+SGD, we follow \cite{liu2018darts} to use initial learning rate 0.025, cosine annealing scheduler, momentum 0.9 and weight decay $3\times10^{-4}$. For DARTS+Adas, we use initial learning rate 0.175, scheduler beta 0.98, momentum 0.9 and weight decay $3\times10^{-4}$. As for architecture parameter optimization, we use the same optimizer and hyper-parameters as in \cite{liu2018darts}. 

% In ADP experiments, most hyperparameters are the same except that we use a batch size of 32. We also increase the initial learning rate of DARTS+SGD to 0.175 for model weights optimization.

% \subsubsection{Architecture Evaluation}
% The searched network is discretized and then trained from scratch for final evaluation. In both CIFAR and CPath experiments, we follow \cite{liu2018darts} to train the network for 600 epochs with batch size 96 and initial channels 36. We use SGD optimizer with an initial learning rate of 0.025, cosine annealing scheduler, momentum 0.9, and weight decay $3\times10^{-4}$. Additional enhancements include cutout and auxiliary towers as in \cite{liu2018darts}.

\subsection{Results on the CIFAR Dataset}
On the CIFAR dataset, we search for the optimum optimizer and number of intermediate nodes. During searching, eight cells are stacked as in DARTS, while the number of nodes in each cell is tuned between 4, 5, and 6. During evaluation, to prevent the network size from being too large with more nodes, we also change the number of cells accordingly, \ie, 20 cells for 4 nodes, 17 cells for 5 nodes, and 14 cells for 6 nodes. Table.~\ref{tab:node_cifar10} and Table.~\ref{tab:node_cifar100} show the test performance of architectures searched on CIFAR-10 and CIFAR-100. We can see that in each setting, DARTS+Adas outperforms the default DARTS+SGD in terms of test accuracy, while a cost is paid in parameter size. This is because the original DARTS+SGD tends to select skip-connections in the final architecture as described above in Sec.~\ref{subsec:metric}. The optimum number of nodes is 4.

\begin{table}[htp]
    \setlength\tabcolsep{1pt} 
    \center
    \caption{Performance of architectures searched with different number of intermediate nodes on CIFAR-10 (32x32 image resolution).}
	\label{tab:node_cifar10}
    \scriptsize{
    \begin{tabular}{ c||c|c|c||c|c|c} %||c|c|c  }
        \hlinewd{1pt}
        \multirow{2}{*}{\textbf{Number of nodes}} &
        \multicolumn{3}{c||}{\textbf{DARTS+SGD}} &
        \multicolumn{3}{c}{\textbf{DARTS+Adas}} \\ % &
        %\multicolumn{3}{c}{\textbf{DARTS+Adas+Stop}} \\
        \cline{2-7}
        & Acc (\%) & Prm (M) & MAC (G) & Acc (\%) & Prm (M) & MAC (G)\\% & Acc (\%) & Prm (M) & Mac (G) \\
        \hline
        \hline
        4 & 97.24\textsubscript{0.09} & 3.30 & 0.55 & 97.43\textsubscript{0.05} & 4.14 & 0.66 \\
        5 & 96.69\textsubscript{0.07} & 3.46 & 0.57 & 97.45\textsubscript{0.06} & 4.23 & 0.67 \\
        6 & 96.16\textsubscript{0.05} & 2.02 & 0.34 & 96.73\textsubscript{0.07} & 4.35 & 0.69 \\
        \hlinewd{1pt}
    \end{tabular}
    }
\end{table}

\begin{table}[htp]
    \setlength\tabcolsep{1pt} 
    \center
    \caption{Performance of architectures searched with different number of intermediate nodes on CIFAR-100 (32x32 image resolution).}
	\label{tab:node_cifar100}
    \scriptsize{
    \begin{tabular}{ c||c|c|c||c|c|c} %||c|c|c  }
        \hlinewd{1pt}
        \multirow{2}{*}{\textbf{Number of nodes}} &
        \multicolumn{3}{c||}{\textbf{DARTS+SGD}} &
        \multicolumn{3}{c}{\textbf{DARTS+Adas}} \\ % &
        %\multicolumn{3}{c}{\textbf{DARTS+Adas+Stop}} \\
        \cline{2-7}
        & Acc (\%) & Prm (M) & MAC (G) & Acc (\%) & Prm (M) & MAC (G)\\% & Acc (\%) & Prm (M) & Mac (G) \\
        \hline
        \hline
        4 & 81.09\textsubscript{0.24} & 2.42 & 0.38 & 84.09\textsubscript{0.18} & 4.24 & 0.68 \\
        5 & 82.10\textsubscript{0.25} & 2.92 & 0.46 & 83.49\textsubscript{0.23} & 4.41 & 0.70 \\
        6 & 81.81\textsubscript{0.24} & 3.25 & 0.52 & 83.15\textsubscript{0.27} & 4.11 & 0.66 \\
        \hlinewd{1pt}
    \end{tabular}
    }
\end{table}

\subsection{Results on the ADP Dataset}
\label{subsec:results_adp}

\begin{table}[htp]
    \setlength\tabcolsep{1pt} 
    \center
    \caption{\label{tab:num_cell}Performance of architectures searched with different optimization strategies and different number of cells on ADP (272x272 image resolution).}
    \scriptsize{
    \begin{tabular}{ c||c|c|c||c|c|c} %||c|c|c  }
        \hlinewd{1pt}
        \multirow{2}{*}{\textbf{Number of cells}} &
        \multicolumn{3}{c||}{\textbf{DARTS+SGD}} &
        \multicolumn{3}{c}{\textbf{DARTS+Adas}} \\% &
        %\multicolumn{3}{c}{\textbf{DARTS+Adas+Stop}} \\
        \cline{2-7}
        & Acc (\%) & Prm (M) & MAC (G) & Acc (\%) & Prm (M) & MAC (G) \\ % & Acc (\%) & Prm (M) & Mac (G) \\
        \hline
        \hline
        4 & 94.40\textsubscript{0.07} & 0.47 & 0.35 & 94.44\textsubscript{0.07} & 0.38 & 0.30 \\ %& 94.44 & 0.50 & 0.38\\
        5 & 94.31\textsubscript{0.05} & 0.55 & 0.42 & 94.36\textsubscript{0.06} & 0.47 & 0.43 \\ %& 94.28 & 0.62 & 0.58\\
        6 & 94.23\textsubscript{0.08} & 0.58 & 0.50 & 94.51\textsubscript{0.06} & 0.52 & 0.51 \\ %& 94.33 & 0.66 & 0.59\\
        \hlinewd{1pt}
    \end{tabular}
    }
\end{table}

\begin{table}[htp]
    \setlength\tabcolsep{1pt} 
    \center
    \caption{\label{tab:num_node}Performance of architectures searched with different number of intermediate nodes on ADP (272x272 image resolution).}
	\label{results_node}
    \scriptsize{
    \begin{tabular}{ c||c|c|c||c|c|c} %||c|c|c  }
        \hlinewd{1pt}
        \multirow{2}{*}{\textbf{Number of nodes}} &
        \multicolumn{3}{c||}{\textbf{DARTS+SGD}} &
        \multicolumn{3}{c}{\textbf{DARTS+Adas}} \\ % &
        %\multicolumn{3}{c}{\textbf{DARTS+Adas+Stop}} \\
        \cline{2-7}
        & Acc (\%) & Prm (M) & MAC (G) & Acc (\%) & Prm (M) & MAC (G)\\% & Acc (\%) & Prm (M) & Mac (G) \\
        \hline
        \hline
        2 & 94.28\textsubscript{0.06} & 0.24 & 0.20 & 94.39\textsubscript{0.02} & 0.24 & 0.21 \\
        3 & 94.36\textsubscript{0.05} & 0.32 & 0.26 & 94.46\textsubscript{0.03} & 0.31 & 0.27 \\
        4 & 94.40\textsubscript{0.07} & 0.47 & 0.35 & 94.44\textsubscript{0.07} & 0.38 & 0.30 \\
        5 & 94.43\textsubscript{0.08} & 0.60 & 0.43 & 94.29\textsubscript{0.05} & 0.77 & 0.52 \\
        6 & 94.26\textsubscript{0.05} & 0.87 & 0.57 & 94.23\textsubscript{0.04} & 0.85 & 0.57 \\
        \hlinewd{1pt}
    \end{tabular}
    }
\end{table}

\begin{figure}[htp]
\centering
\includegraphics[width=0.47\textwidth]{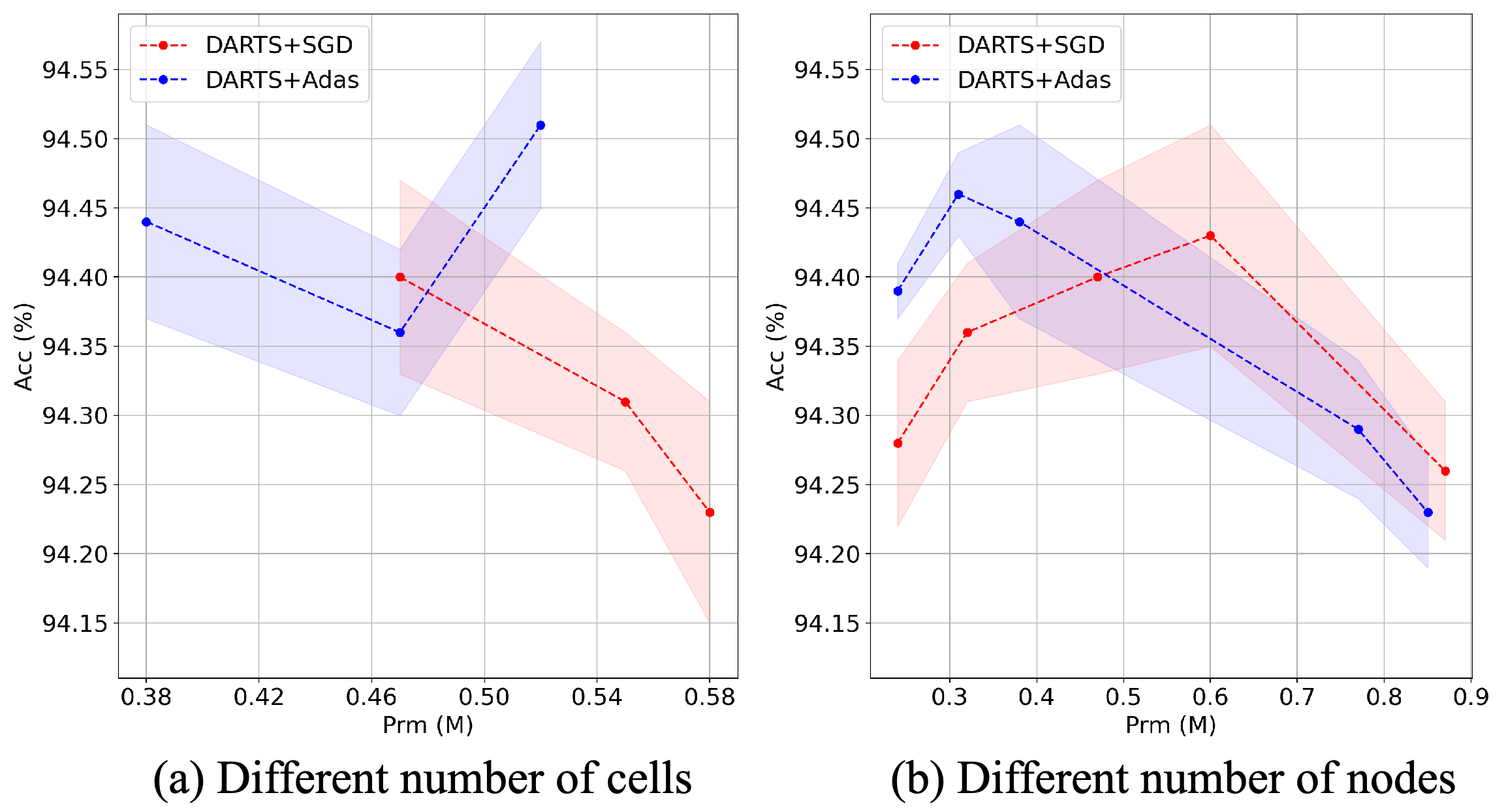}
\caption{(a) Performance of different numbers of cells. Each dot denotes a different cell number in 4, 5, and 6. (b) Performance of different numbers of nodes. Each dot represents a different node number in 2, 3, 4, 5, and 6.}
\label{fig:cell_node_test}
\end{figure}

\begin{figure*}[htp]
\centering
\includegraphics[width=0.85\textwidth]{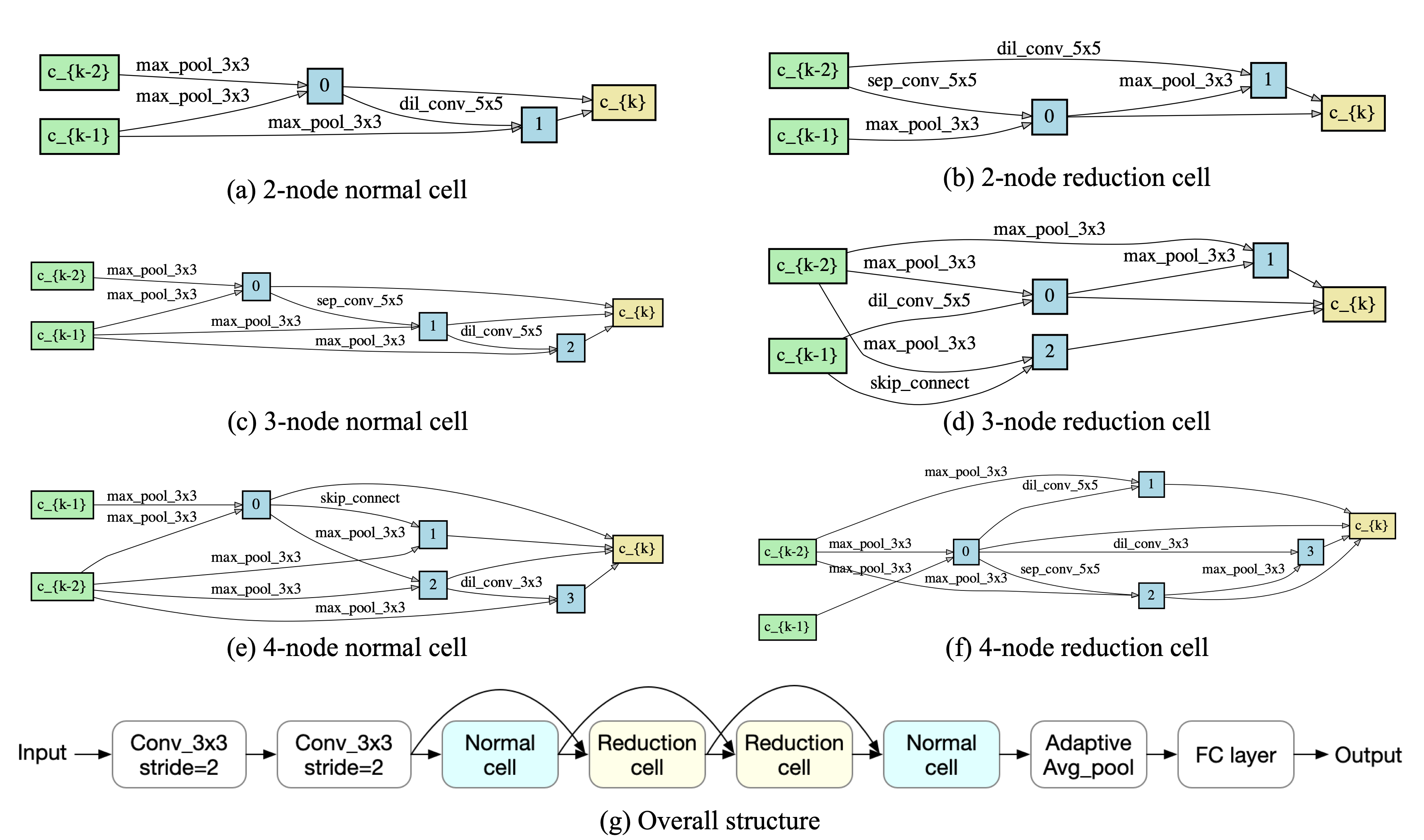}
% \subfigure[2-node normal]{\includegraphics[width=0.41\textwidth]{figures/normal_n_2_c_4.eps}}
% \subfigure[2-node reduction]{\includegraphics[width=0.41\textwidth]{figures/reduction_n_2_c_4.eps}}
% \subfigure[3-node normal]{\includegraphics[width=0.41\textwidth]{figures/normal_n_3_c_4.eps}}
% \subfigure[3-node reduction]{\includegraphics[width=0.41\textwidth]{figures/reduction_n_3_c_4.eps}}
% \subfigure[4-node normal]{\includegraphics[width=0.41\textwidth]{figures/best_normal_n_4_c_4.eps}}
% \subfigure[4-node reduction]{\includegraphics[width=0.41\textwidth]{figures/best_reduction_n_4_c_4.eps}}
% \subfigure[Overall architecture]{\includegraphics[width=0.8\textwidth]{figures/network_macro_structure.png}}
\caption{(a)-(f) Best cell structures found in different cases. (g) Overall network structure.}
\label{fig:snapshots_nodes}
\end{figure*}

To find the optimum architecture on the ADP dataset, we run the architecture search in all different parameter settings, \ie, different numbers of cells and nodes, different choices of optimizer. Note that in CIFAR experiments we search for a shallower network (with few cells) during searching but train a deeper one (with more cells) for evaluation due to the complexity of CV datasets. In ADP, however, we keep the number of cells the same in two stages. This is because ADP is a simpler dataset so we don't need to increase the model complexity during evaluation. This also brings more consistency to the search-evaluation pipeline. 
% In each setting, we conduct the search four times with different random seeds and select the best architecture based on their performance in a quick evaluation (training from scratch for 100 epochs). Then a final evaluation is performed on this selected architecture. 

\textbf{Optimum optimizer and number of cells.} We first search for the optimum optimizer and the number of cells. The test results of the searched architectures after final evaluation are shown in Table.~\ref{tab:num_cell}. We also plot the accuracy versus parameter size in Fig.~\ref{fig:cell_node_test} (a), where each dot represents a different choice of cell number. We can see that as the number of cells increases the accuracy of DARTS+SGD drops while its parameter size increases. When using DARTS+Adas, the test accuracy remains the highest across different numbers of cells, and the parameter size remains the smallest. The highest accuracy is achieved with 6 cells, while for 4 cells, the searched architecture has the smallest size but still obtains the second highest accuracy.

\textbf{Optimum number of intermediate nodes.} 
We then fix the number of cells as four and search for the optimum number of intermediate nodes. The test performance are shown in Tabel.\ref{tab:num_node} and in Fig.~\ref{fig:cell_node_test} (b). We can see that DARTS+Adas achieves higher accuracy than DARTS+SGD with fewer nodes, hence less computation complexity. The highest accuracy $94.46\%$ is achieved with 3 nodes, leading to 0.31M parameters and 0.27G MAC operations. Fig.~\ref{fig:snapshots_nodes} shows the cell architectures searched with 2, 3, and 4 nodes using DARTS+Adas.

\subsection{Architecture Transferabilty}
\label{subsec:transfer}

\begin{table*}[htp]
    \setlength\tabcolsep{1pt} 
    \center
    \caption{\label{tab:transfer}Performance of different networks on different datasets. Note that {\color{darkgreen}\textbf{green}} indicates the best, and {\color{orange}\textbf{orange}} indicates within standard deviation from the best.}
    \scriptsize{
    \begin{tabular}{ c||c|c|c||c|c|c||c|c|c||c|c|c}
        \hlinewd{1pt}
        \multirow{2}{*}{\textbf{Network}}&
        \multicolumn{3}{c||}{\textbf{ADP} \cite{hosseini2019atlas}} &
        \multicolumn{3}{c||}{\textbf{BCSS} \cite{BCSSpaper}} &
        \multicolumn{3}{c||}{\textbf{BACH} \cite{aresta2019bach}} &
        \multicolumn{3}{c}{\textbf{Osteosarcoma} \cite{leavey2019osteosarcoma}} \\
        \cline{2-13}
        & Acc (\%) & Prm (M) & MAC (G) & Acc (\%) & Prm (M) & MAC (G) & Acc (\%) & Prm (M) & MAC (G) & Acc (\%) & Prm (M) & MAC (G)\\
        \hline
        \hline
        ResNet18 \cite{he2016deep}& 93.43\textsubscript{0.35} & 11.19 & 2.76 &  96.13\textsubscript{0.80} & 11.18 & 2.76 & 90.05\textsubscript{0.89} & 11.18 & 2.76 & 93.23\textsubscript{0.30} & 11.18 & 2.76\\
        % \hline
        % ResNeXt50 \cite{xie2017aggregated} & 93.94 & 23.05 & 6.42 &  & & & & & & & &\\
        % \hline
        MobileNetV2 1.0 \cite{sandler2018mobilenetv2} & 93.16\textsubscript{0.22} & 2.27 & 0.48 & 93.90\textsubscript{2.11} & 2.24 & 0.48 & 89.55\textsubscript{0.95} & 2.23 & 0.48 & 90.73\textsubscript{0.46} & 2.23 & 0.48\\
        % \hline
        MobileNetV2 0.35 \cite{sandler2018mobilenetv2} & 91.73\textsubscript{0.79} & 0.44 & 0.10 & 93.84\textsubscript{1.02} & 0.41 & 0.10 & 87.18\textsubscript{0.46} & 0.40 & 0.10 & 90.59\textsubscript{0.50} & 0.40 & 0.10\\
        % \hline
        MobileNetV3-large \cite{howard2019searching} & 91.43\textsubscript{0.82} & 4.24 & 0.34 & 94.40\textsubscript{0.62} & 4.21 & 0.34 & 86.98\textsubscript{1.61} & 4.21 & 0.34 & 89.60\textsubscript{0.69} & 4.21 & 0.34\\
        % \hline 
        MobileNetV3-small \cite{howard2019searching} & 92.15\textsubscript{0.21} & 1.55 & \color{darkgreen}\textbf{0.09} & 93.70\textsubscript{0.39} & 1.53 & \color{darkgreen}\textbf{{0.09}} & 86.32\textsubscript{1.16} & 1.52 & \color{darkgreen}\textbf{0.09} & 89.11\textsubscript{0.89} & 1.52 & \color{darkgreen}\textbf{0.09}\\
        % \hline
        ShuffleNetV2 1.0 \cite{ma2018shufflenet} & 92.18\textsubscript{1.62} & 1.29 & 0.23 & 96.14\textsubscript{0.73} & 1.26 & 0.23 & 91.41\textsubscript{0.69} & 1.26 & 0.23 & 91.67\textsubscript{0.57} & 1.26 & 0.23\\
        % \hline
        SENet18 \cite{hu2018squeeze} & 93.33\textsubscript{0.18} & 11.28 & 2.76 & 96.63\textsubscript{0.27} & 11.27 & 2.76 & 90.93\textsubscript{0.78} & 11.27 & 2.76 & 92.10\textsubscript{1.14} & 11.27 & 2.76\\
        % \hline
        MNASNet-A1 \cite{tan2019mnasnet} & 91.98\textsubscript{0.21} & 2.65 & 0.50 & 95.84\textsubscript{0.11} & 2.62 & 0.50 & 83.63\textsubscript{2.40} & 2.61 & 0.50 & 90.26\textsubscript{0.64} & 2.61 & 0.50\\
        % \hline
        MNASNet-small \cite{tan2019mnasnet} & 92.44\textsubscript{0.17} & 0.79 & 0.11 & 94.63\textsubscript{1.51} & 0.76 & 0.11 & 83.54\textsubscript{2.15} & 0.75 & 0.11 & 91.47\textsubscript{0.64} & 0.75 & 0.11\\
        % \hline
        DARTS 4-cell\cite{liu2018darts}& 94.24\textsubscript{0.05} & 0.49 & 0.38 & \color{orange}\textbf{97.38\textsubscript{0.03}} & 0.48 & 0.38 & \color{darkgreen}\textbf{93.77\textsubscript{0.26}} & 0.48 & 0.38 & \color{darkgreen}\textbf{95.04\textsubscript{0.27}} & 0.48 & 0.38 \\
        \hline
        DARTS-ADP-N4 & \color{orange}\textbf{94.37\textsubscript{0.00}} & 0.38 & 0.30 & \color{darkgreen}\textbf{97.39\textsubscript{0.02}} & 0.36& 0.30 & \color{orange}\textbf{93.74\textsubscript{0.18}} & 0.36 & 0.30 & 93.99\textsubscript{0.06} & 0.36 & 0.30\\
        % \hline
        DARTS-ADP-N3 & \color{darkgreen}\textbf{94.41\textsubscript{0.04}} & 0.31 & 0.27 & 97.34\textsubscript{0.07} & 0.30 & 0.27 & 92.07\textsubscript{2.15} & 0.30 & 0.27 & 94.52\textsubscript{0.24} & 0.30 & 0.27\\
        % \hline
        DARTS-ADP-N2 & 94.34\textsubscript{0.03} & \color{darkgreen}\textbf{0.24} & 0.21 & \color{orange}\textbf{97.38\textsubscript{0.03}} & \color{darkgreen}\textbf{0.24} & 0.21 & 93.38\textsubscript{1.21} & \color{darkgreen}\textbf{0.23} & 0.21 & 94.54\textsubscript{0.35} & \color{darkgreen}\textbf{0.23} & 0.21\\
        \hlinewd{1pt}
    \end{tabular}
    }
\end{table*}

\begin{figure*}[htp!]
\centering
\subfigure[ADP]{\includegraphics[width=0.25\textwidth]{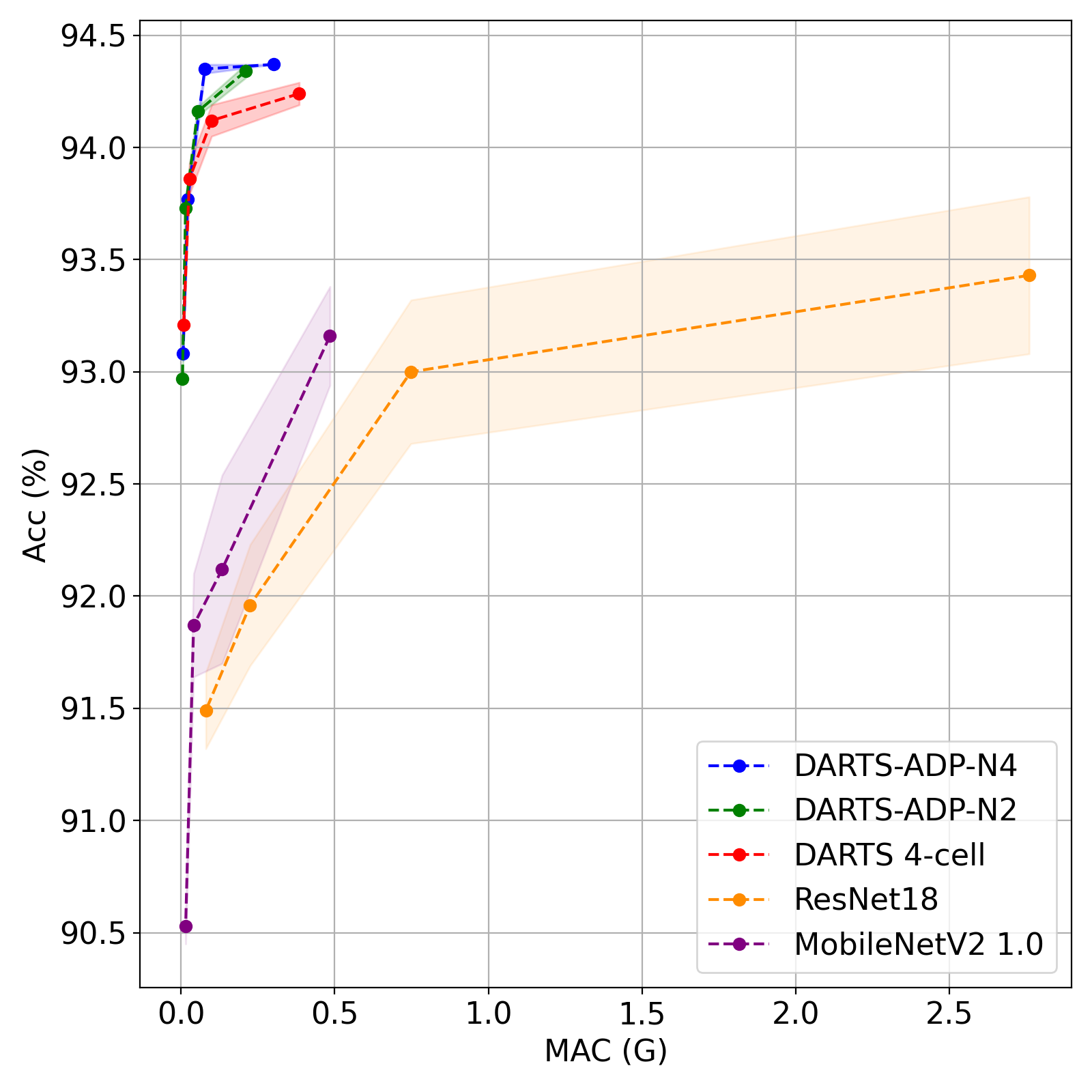}}
\subfigure[BCSS]{\includegraphics[width=0.25\textwidth]{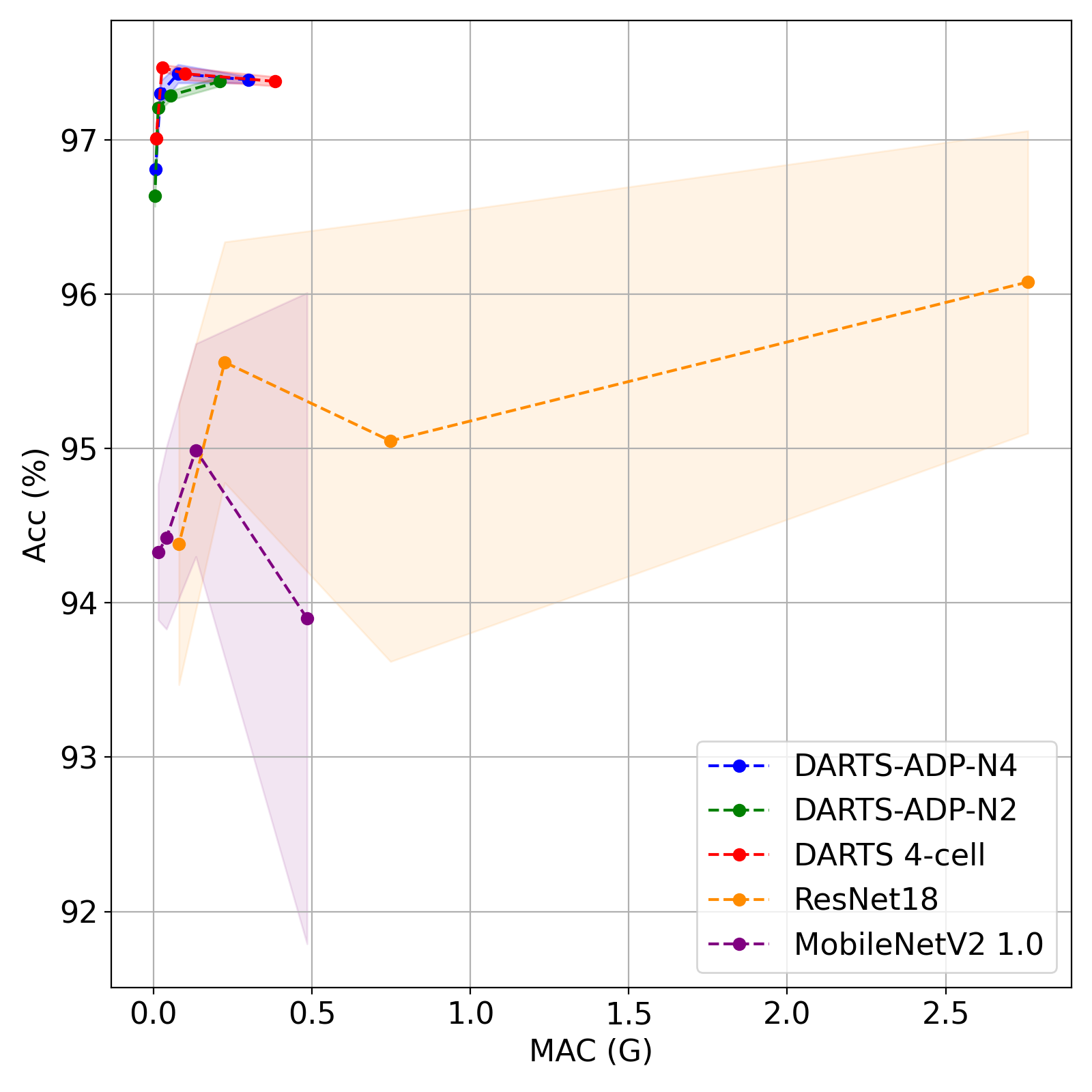}}
\subfigure[BACH]{\includegraphics[width=0.25\textwidth]{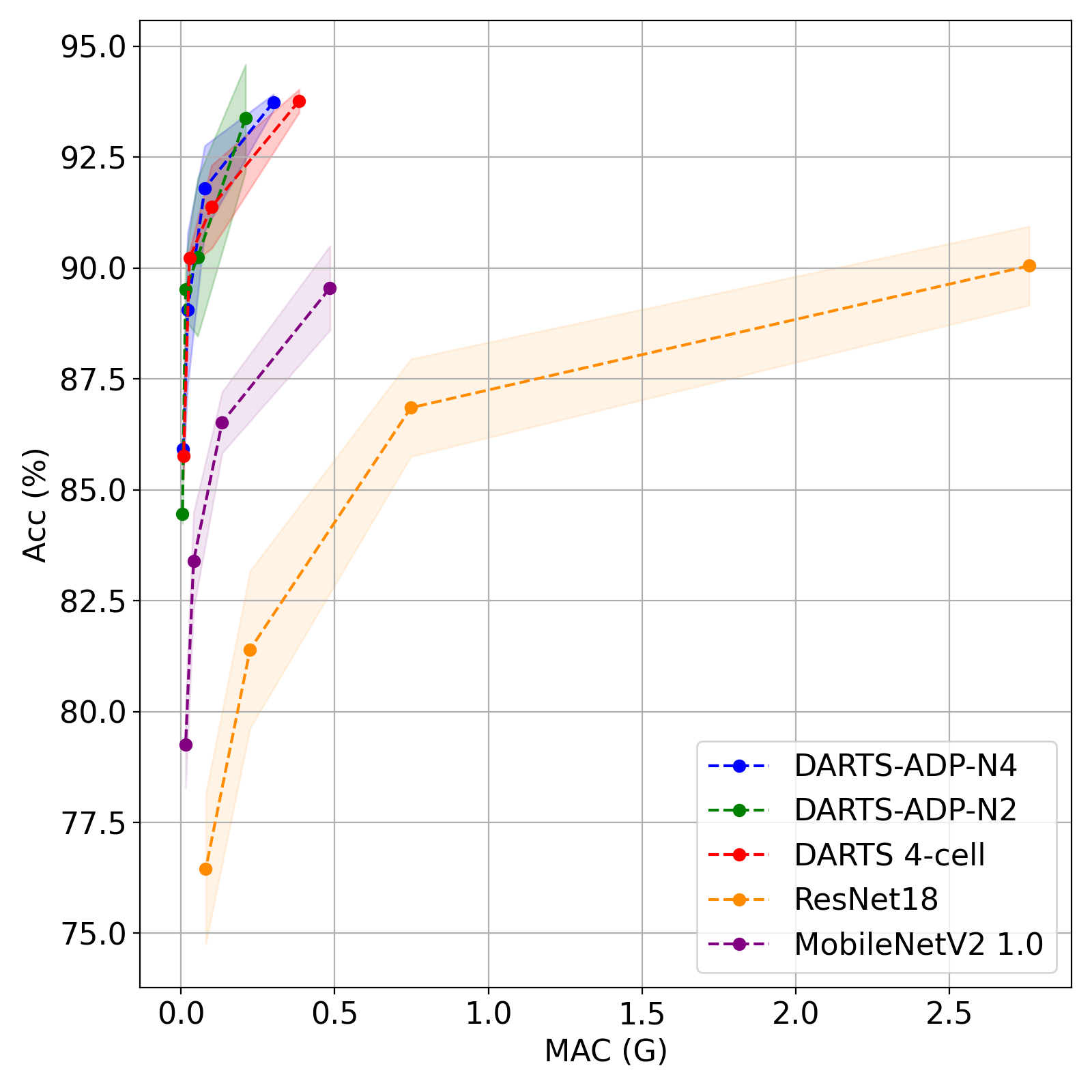}}
\caption{Performance of different image resolution (denoted by different dots) on different dataset and network selection.}
\label{fig:resolution}
\end{figure*}

We select the searched architectures with 2, 3 and 4 nodes (namely DARTS-ADP-N2, -N3 and -N4), and train them on three more datasets: BCSS \cite{BCSSpaper}, BACH \cite{aresta2019bach} and Osteosarcoma \cite{leavey2019osteosarcoma}. Details including data augmentations can be found in Section Datasets of the Supplementary Material. The goal is to evaluate how the searched architectures perform when transferred to different CPath datasets that cover different variations of single- vs multi-labels, multiclass problems, data samples, and organs. We also train several mobile-friendly architectures on them for comparison, including a 4-cell DARTS~\cite{liu2018darts}. All networks are trained for 600 epochs with batch size 96, using the SGD optimizer with a 0.025 initial learning rate and cosine annealing scheduler. 
% We also take the cell architecture searched in DARTS \cite{liu2018darts} and build a 4-cell network (DARTS 4-cell) for evaluation.

As shown in Tabel.\ref{tab:transfer}, across all datasets, the group of DARTS-ADP networks achieves higher or comparable test accuracy than state-of-the-art networks, but with smaller parameter sizes. The 2-node version contains only 0.24M parameters but still ranks high in test accuracy. As for computation complexity, though MobileNetV2 0.35 \cite{sandler2018mobilenetv2}, MobileNetV3-small \cite{howard2019searching}, and MNASNet-small \cite{tan2019mnasnet} achieve fewer MAC operations, their accuracies are two percent lower. This shows the superiority of DARTS-ADP in the speed-accuracy trade-off, which is desirable for CPath applications of high-throughput image analysis.

\subsection{Performance of different image resolution}
\label{subsec:resolution}

Another way to meet the needs of high-throughput applications is to decrease the image resolution. To evaluate the robustness of the architectures against downscaled inputs, we retrain several models with different resolutions (272, 136, 68, 34) in three datasets. Fig.~\ref{fig:resolution} shows the performance of different network selection. Each line represents a network and each dot represents a specific resolution. The DARTS-based networks consistently achieve the highest accuracy with the lowest computation complexity in all resolutions and across all datasets. As the resolution decreases, their test accuracy exhibits a much less drop compared to ResNet18 \cite{he2016deep} and MobileNetV2 \cite{sandler2018mobilenetv2}, which shows the robustness of the DARTS-based networks. Such robustness is also illustrated in the standard deviation (denoted by shades). Compared to 4-cell DARTS \cite{liu2018darts}, the two DARTS-ADP networks obtain higher or comparable test accuracy with lower computation complexity, which again shows their superiority in the speed-accuracy trade-off. 

\subsection{Grad-CAM analysis}
\label{subsec:gradcam}

\begin{figure}[htp]
\centering
\subfigure[ADP]{\label{fig:gradcam_adp}\includegraphics[width=0.35\paperwidth]{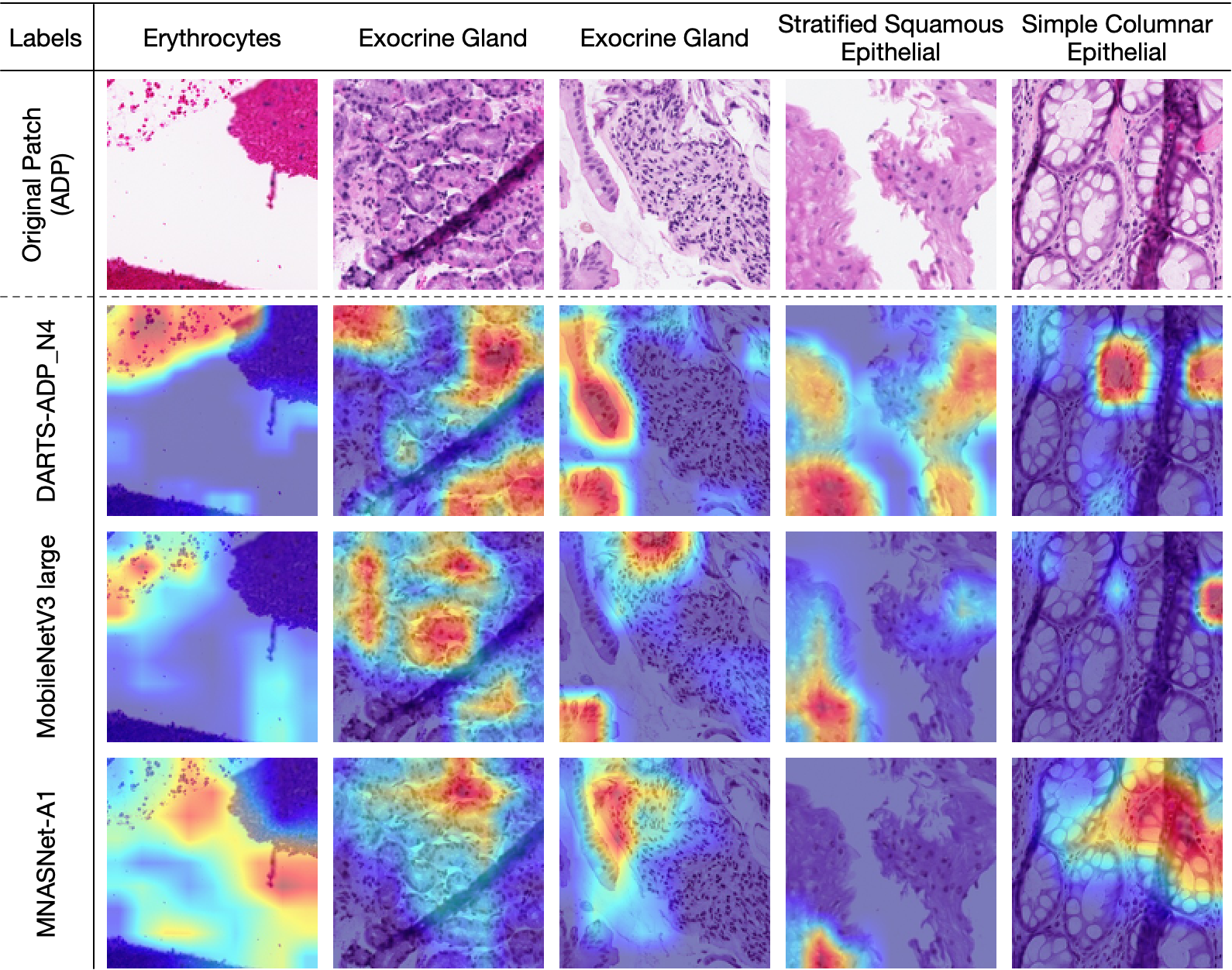}}
\subfigure[BCSS]{\label{fig:gradcam_bcss}\includegraphics[width=0.35\paperwidth]{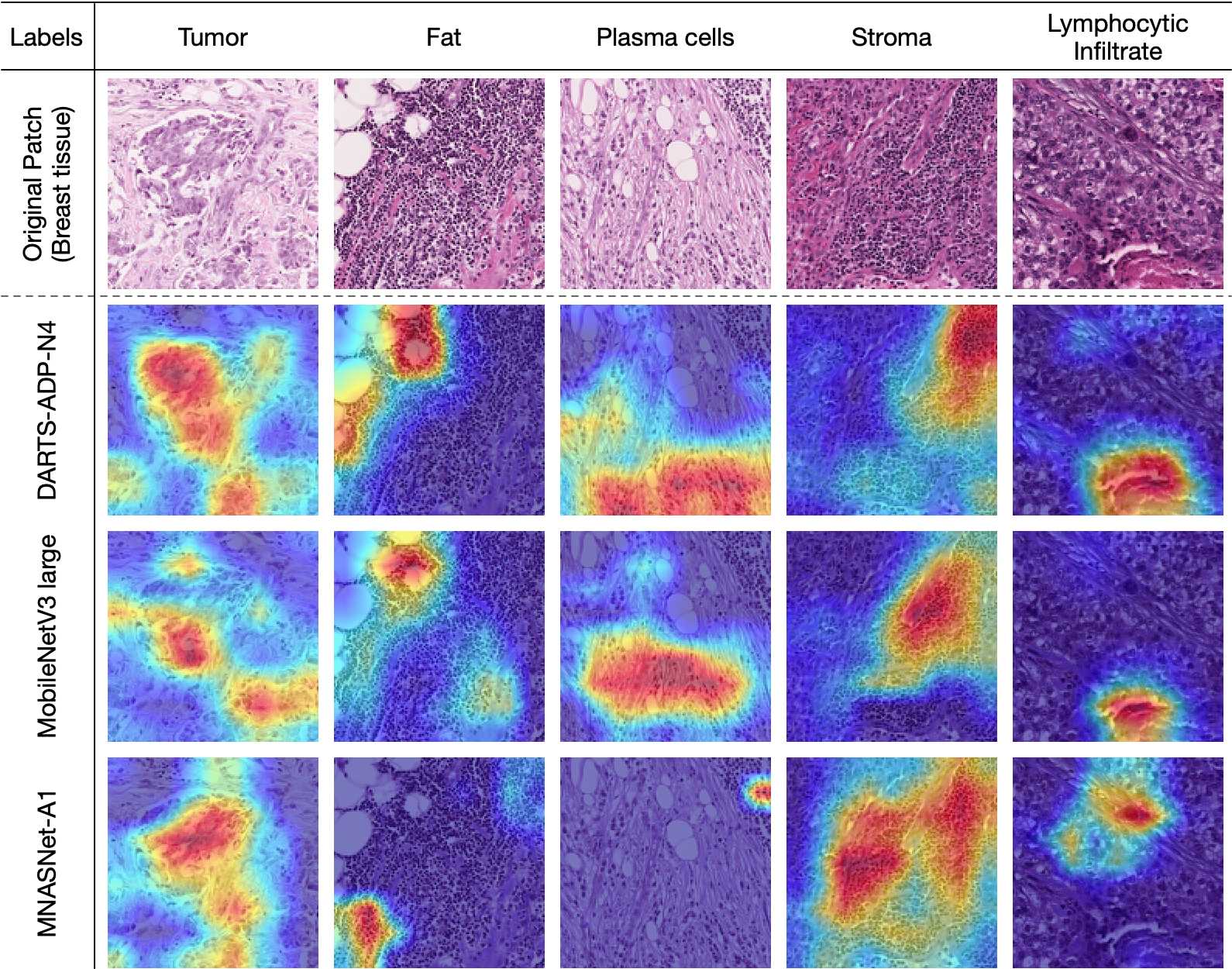}}
\subfigure[BACH]{\label{fig:gradcam_bach}\includegraphics[width=0.35\paperwidth]{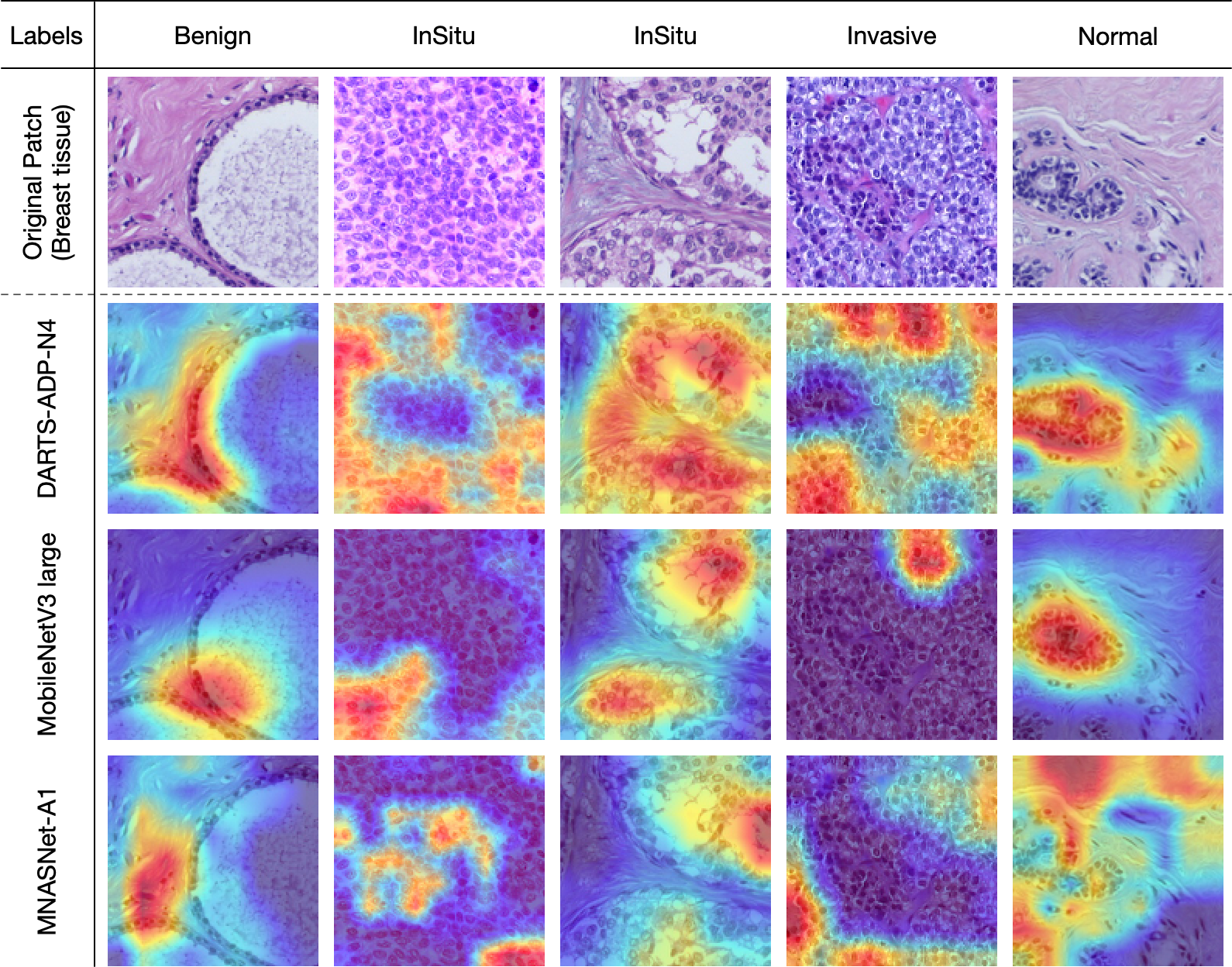}}
\caption{Grad-CAM performance of different network selection on different datasets.}
\label{fig:gradcam}
\end{figure}

To better understand and interpret the performance of feature representation of different networks, we apply Grad-CAM \cite{selvaraju2017grad} on the last convolutional layers of different networks to obtain the heatmaps for predicting ground truth labels. This visualization technique allows us to evaluate the reliability of different networks in label prediction. We randomly select five patches that contain different ground truth labels from the test set of ADP, BCSS, and BACH, and feed them into three networks for comparison. The selected networks are DARTS-ADP-N4, MobileNetV3-large \cite{howard2019searching}, and MNASNet-A1 \cite{tan2019mnasnet}. Results are shown in Fig.~\ref{fig:gradcam}, where the heatmap indicates pixel-level confidence of pertinent labels of the image patch. 

According to pathologists' assessment, the overall performance of DARTS-ADP-N4 is the best. Examples are shown in the first column of Fig.~\ref{fig:gradcam_adp}, where DARTS-ADP-N4 successfully highlights the region of \textit{Erythrocytes}. Either MobileNetV3-large or MNASNet-A1 discovers incomplete or false regions. This demonstrates the superiority of DARTS-ADP in extracting label-pertinent features from image patches, and hence more reliable predictions.

%------------------------------------------------------------------------
\section{Conclusion}

In this paper, we propose a general DARTS-based searching framework for CPath applications. We first use a probing metric to show that the existing DARTS lacks proper hyperparameter tuning, and how the generalization performance of the searched model can be improved with an adaptive optimization strategy. We then apply this searching framework on a histological tissue type dataset ADP and develop architectures that outperform the state-of-the-art networks with higher prediction accuracy and lower computation complexity. We transfer the searched architectures to other CPath datasets including BCSS, BACH, and Osteosarcoma, and conduct extensive experiments to demonstrate the robustness and reliability of the networks in various cases.

\clearpage
\twocolumn[{
\begin{center}
	{\Large \bf Supplementary Material for ``Probeable DARTS with Application to Computational Pathology''}
\end{center}
}]
\setcounter{section}{0}
\makeatletter

\section{Network Structures}
\begin{figure}[htp]
    \centering
    \includegraphics[width=0.35\textwidth]{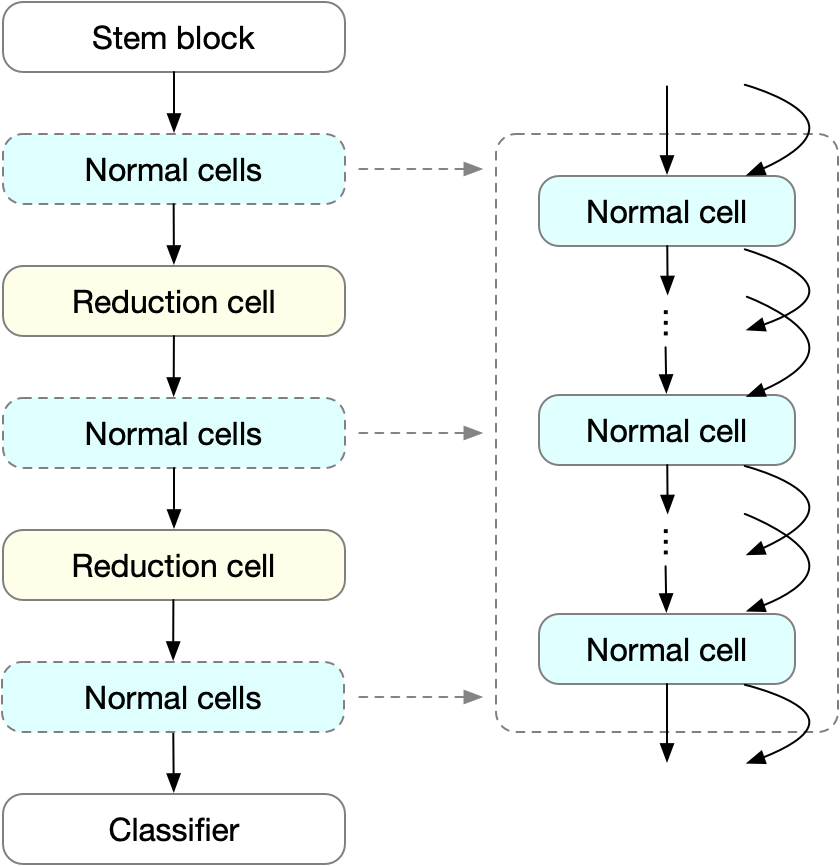}
    \caption{General network structure for searching and evaluation.}
    \label{fig:general_network}
\end{figure}

The macro network structures in both the searching and evaluation phases are formed by stacking the normal and reduction cells sequentially. At $1/3$ and $2/3$ of the total depth of the network, there are reduction cells. Fig.~\ref{fig:general_network} shows the general network structure, where the stem block contains several convolutional layers and the classifier consists of a global pooling layer and a fully connected layer. 

\begin{figure}[htp]
    \centering
    \includegraphics[width=0.15\textwidth]{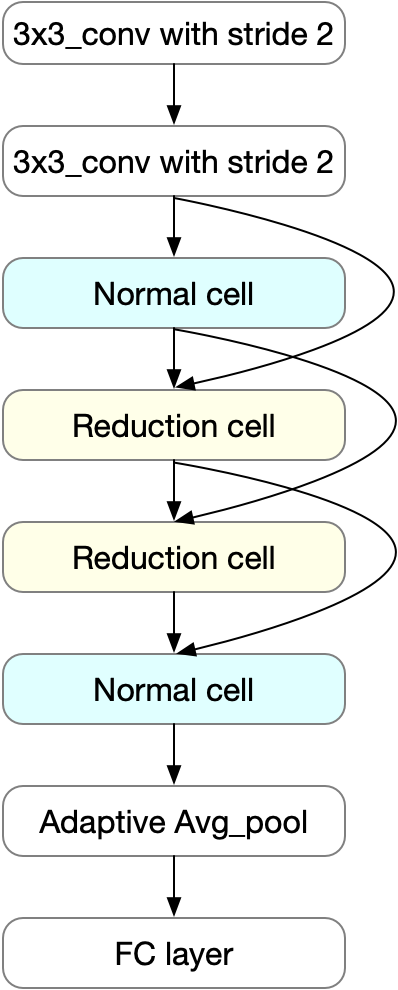}
    \caption{Final network structure searched on ADP.}
    \label{fig:adp_network}
\end{figure}

The final architecture searched on ADP \cite{hosseini2019atlas} is shown in Fig.~\ref{fig:adp_network}. Note that there are no normal cells between the two reduction cells since the total number of cells is four, which is not divisible by three. 

\section{Dataset Details}
\textbf{CIFAR \cite{krizhevsky2009learning}.} In the searching phase, we follow \cite{liu2018darts} to split the original training set into two parts, one for training and one for evaluation. In the evaluation phase, we use the default splits. We use random cropping with size 32x32 and random horizontal flipping as data augmentations.

\textbf{CPath datasets.} ADP and BCSS \cite{BCSSpaper} are multi-label datasets, while BACH \cite{aresta2019bach} and Osteosarcoma \cite{leavey2019osteosarcoma} are single-label. Their image resolution is all 272x272. We only conduct searching on ADP but evaluate the searched architecture on all four datasets. During searching, we treat half of the training set of ADP as the validation set. Data augmentations in all datasets include random horizontal and vertical flipping, random affine, and resize. Note that during searching on ADP, we resize the images to 64x64 to alleviate the computation overhead, and during evaluation, images are resized only in the test of different resolutions (136, 68, and 34).

% You must include your signed IEEE copyright release form when you submit
% your finished paper. We MUST have this form before your paper can be
% published in the proceedings.
\section{Hyperparameters}
\subsection{Architecture Search} 
In CIFAR experiments, we train the network for 50 epochs with batch size 64 and initial channels 16. We test two optimizers for optimizing model weights, which are the original SGD \cite{liu2018darts} and Adas \cite{hosseini2020adas}. For DARTS+SGD, we follow \cite{liu2018darts} to use initial learning rate 0.025, cosine annealing scheduler, momentum 0.9 and weight decay $3\times10^{-4}$. For DARTS+Adas, we use initial learning rate 0.175, scheduler beta 0.98, momentum 0.9, and weight decay $3\times10^{-4}$. As for architecture parameter optimization, we follow \cite{liu2018darts} to use Adam \cite{kingma2014adam} optimizer with initial learning rate $3\times10^{-4}$, momentum $(0.5, 0.999)$, and weight decay $10^{-3}$. 

In ADP experiments, most hyperparameters are the same except that we use batch size 32 due to computation overhead. We also increase the initial learning rate of DARTS+SGD to 0.175 for model weights optimization.

\subsection{Architecture Evaluation}
In both CIFAR and CPath experiments, we follow \cite{liu2018darts} to train the network for 600 epochs with batch size 96 and initial channels 36. We use SGD optimizer with an initial learning rate of 0.025, cosine annealing scheduler, momentum 0.9, and weight decay $3\times10^{-4}$. Additional enhancements include cutout and auxiliary towers as in \cite{liu2018darts}. Note that we disable auxiliary towers in training when we compare the performance of the searched architectures with the state-of-the-art networks.

\end{document}